\documentclass[runningheads]{llncs}

\usepackage{eccv}

\usepackage{eccvabbrv}

\usepackage{graphicx}
\usepackage{booktabs}

\usepackage[accsupp]{axessibility}  

\usepackage{hyperref}
\usepackage{xcolor}
\newcommand{\hlv}[1]{\textcolor{blue}{#1}}
\usepackage{orcidlink}
\usepackage{algorithm}
\usepackage{algorithmic}
\usepackage{wrapfig}
\usepackage{array}
\usepackage{booktabs}
\usepackage{multirow}
\usepackage{tikz}
\usepackage{bm}
\usepackage{booktabs}
\usepackage{tabularx}
\usepackage{makecell} 

\usepackage{xspace}

\begin{document}




\title{\textit{StochasT}: Learning with Stochastic Turn Depth for Visual Instruction Tuning}

\titlerunning{StochasT: Stochastic Turn Depth for VIT}



\author{Yuan Qing\inst{1} \and
Chengzhi Mao\inst{2} \and
Boqing Gong\inst{1}}

\authorrunning{Y.~Qing et al.}

\institute{Boston University, Boston, USA \\ \and
Rutgers University, New Brunswick, USA\\
\email{\{ymqing, bgong\}@bu.edu}, \email{chengzhi.mao@rutgers.edu}}

\maketitle

\vspace{-12pt}
\begin{abstract}
    Large Vision-Language Models (LVLMs) rely extensively on Visual Instruction Tuning (VIT) to elicit their multimodal reasoning capabilities. However, we find a discrepancy: VIT often packs multiple language tasks about the same image for conversational, multi-turn training, whereas existing benchmarks evaluate LVLMs in isolated, single-turn scenarios. The models can suffer from visual attention decay and contextual overfitting during multi-turn training, making it hard for them to realize their full potential in the mismatched test phase. To close the gap, we propose learning with Stochastic Turn Depth (StochasT), which stochastically groups language tasks for the same image into clusters of varying sizes (turn depth) while preserving their organic order. Hence, while StochasT draws on Dropout and stochastic depth for ResNets, it does not actually drop anything to maximize the utility of the training data. Furthermore, we introduce a challenging, benchmark-agnostic evaluation mechanism based on the Balanced Latin Square to measure LVLMs' robustness under varying contextual dependencies. Extensive experiments demonstrate that StochasT effectively grants LVLMs strong, harmonized capabilities for both single-turn and multi-turn use cases. Code is available at: \url{https://yuanqing-ai.github.io/StochasT}.
\end{abstract}

\section{Introduction}
\label{sec:intro}

Large Vision-Language Models (LVLMs)~\cite{comanici2025gemini,bai2025qwen3vltechnicalreport,bai2025qwen25vltechnicalreport,wang2025internvl35advancingopensourcemultimodal,singh2025openai,gemmateam2025gemma3technicalreport} have demonstrated remarkable capabilities across various vision applications, spanning Perception (e.g., object detection, segmentation, OCR), Reasoning (e.g., visual question answering, multi-hop reasoning, chart understanding), and Action (e.g., visual instruction following, embodied navigation, GUI control). Standard approaches for building such systems typically integrate a pretrained Large Language Model (LLM)~\cite{grattafiori2024llama,qwen2025qwen25technicalreport} with a pretrained visual encoder, bridged by an alignment module such as an MLP projection layer or a Q-Former~\cite{li2023blip}.

\begin{figure}
    \centering
    \includegraphics[width=1\linewidth]{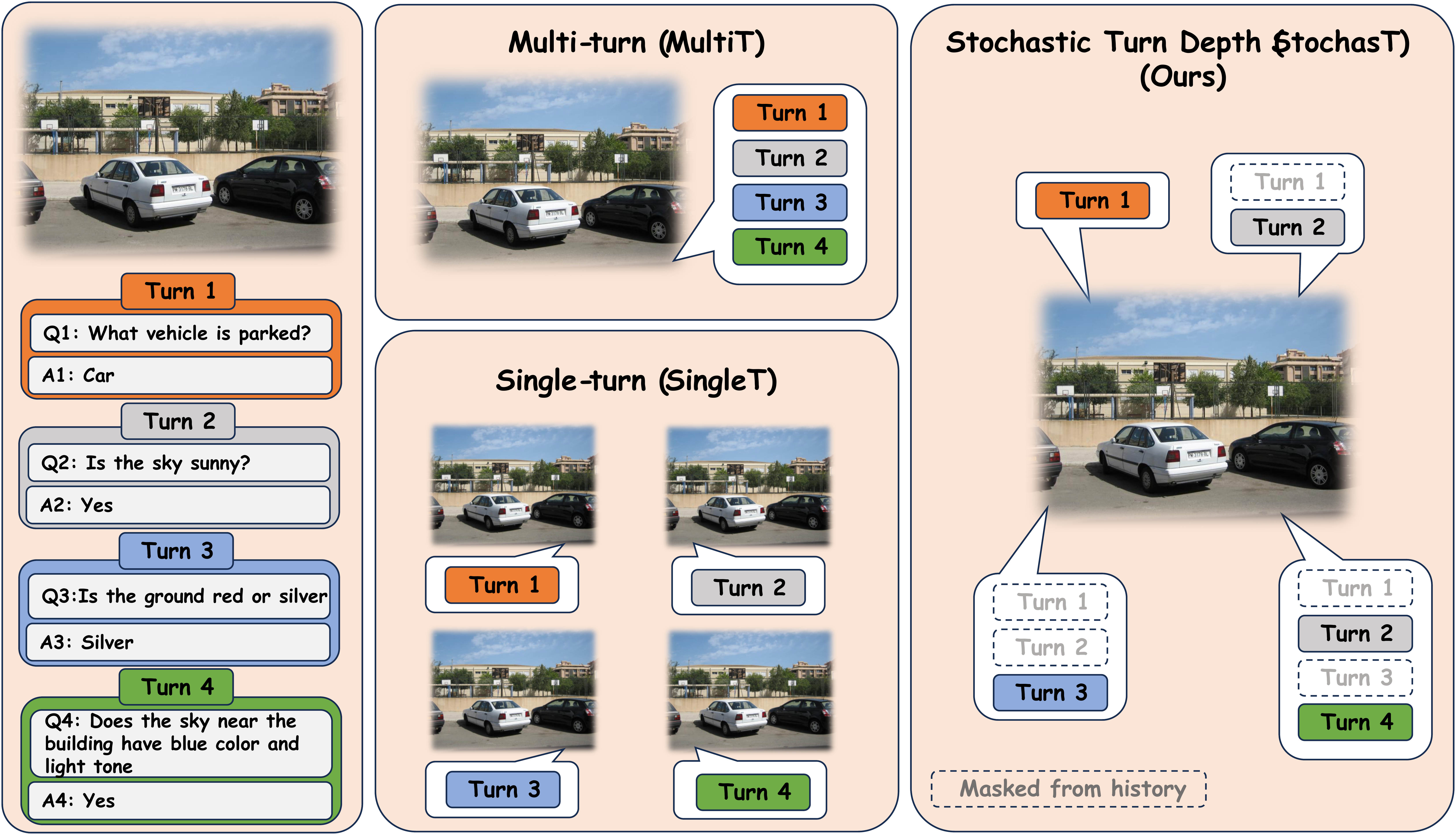}
    \caption{{A visual instruction tuning example (left) and three grouping mechanisms: multi-turn (multiT), singleT, and our proposed stochastic turn depth (\textbf{StochasT}).}}
    \label{fig:teaser1}
\end{figure}

To unlock the multimodal reasoning and generalization potential of LVLMs, Visual Instruction Tuning (VIT)~\cite{liu2023visual,Liu_2024_CVPR,zhu2023minigpt} is critical. Prior studies suggest that much of the world knowledge embedded within foundational LLMs is acquired during large-scale pretraining~\cite{zhou2023lima,tan2025visionllmsbadhierarchical}. Consequently, VIT primarily serves to activate and align this latent knowledge toward downstream multimodal task objectives, rather than learning it from scratch.

\setlength{\intextsep}{4pt}   
\setlength{\columnsep}{8pt}   
\begin{wrapfigure}{r}{0.45\textwidth}
    \centering
    \includegraphics[width=0.48\textwidth]{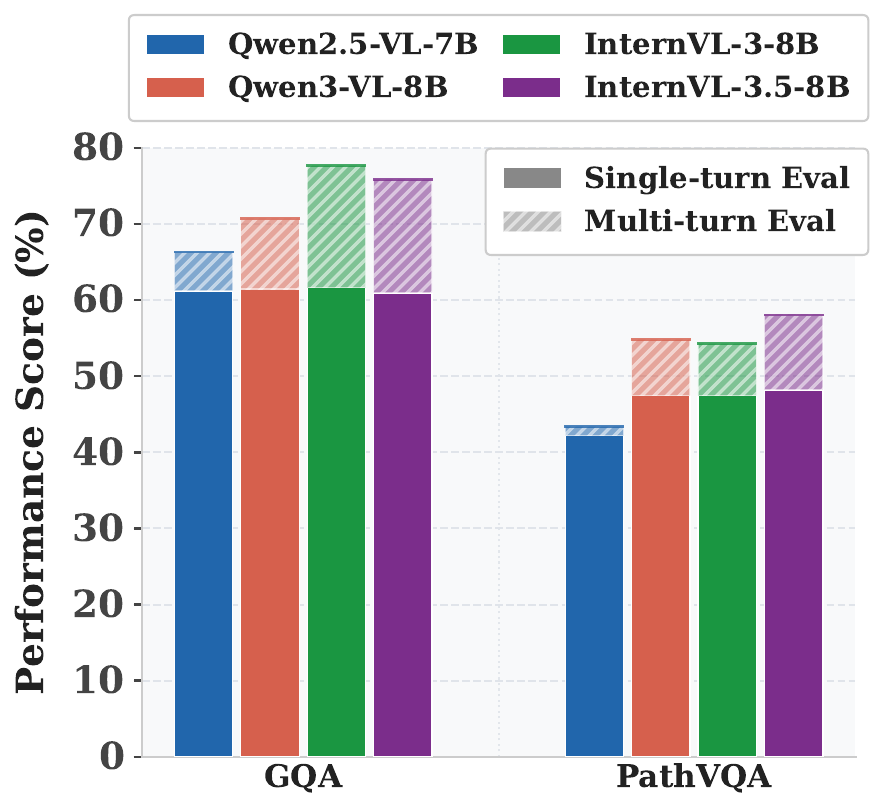}
    \caption{Performance comparison of several SOTA LVLMs under singleT and multiT evaluation.}
    \label{fig:stat}
\end{wrapfigure}
Unlike pure language tasks, the high information density inherent to visual data naturally affords multi-turn (multiT) language queries. A single image often grounds multiple distinct instructions (as illustrated in~\cref{fig:teaser1}), and this one-image-multiT format has been frequently used in VIT~\cite{liu2023visual}. However, a significant discrepancy exists between this multiT training paradigm and currently prevalent single-turn (singleT) evaluation protocols. As depicted in~\cref{fig:teaser1}, multiT training groups all instruction-answer pairs about the same image into one training example, while singleT evaluation anchors every instruction individually to the image. As a result, a model may fail to answer a simple question in isolation but succeed when the exact same question is contextualized within a conversation. 
Most existing LVLM benchmarks~\cite{liu2024mmbench,fu2025mme,li2024seed,yue2024mmmu} employ singleT testing exclusively, treating related questions about the same image as independent, isolated runs. However, if we instead evaluate LVLMs in the same, multiT way as used in training, we can boost their performance significantly (see \cref{fig:stat}; experiment setup in supplement). Surprisingly, the literature has largely overlooked this discrepancy between multiT VIT and singleT evaluation.

In this work, we investigate two primary research questions hinging on this observation: 1) How can we effectively balance the one-image-multiT training paradigm to optimize performance across both singleT and multiT evaluation? 2) Given the substantial discrepancies between singleT and multiT performance, how should we best evaluate LVLMs' robustness in various use cases?




To address the first question, we propose learning with Stochastic Turn Depth (\textbf{StochasT}) to harmonize LVLMs' capabilities for singleT and multiT testing. Inspired by the concept of neuron or layer dropout~\cite{srivastava2014dropout,huang2016deep}, StochasT naturally stochastically varies the number of historical conversational turns for each training sample. Through a controlled expansion of the underlying conversation tree, this strategy substantially enriches the training context distribution and improves contextual diversity. Crucially, this is achieved without introducing additional data or extra computational overhead, effectively training the model on an "ensemble" of multiT data configurations. Furthermore, StochasT preserves the total number of training samples while diversifying their internal structure and depth, allowing it to be seamlessly integrated with other training objectives and novel structures~\cite{zhou2025learning,oh2025visual,wang2025reconstructive,cha2024honeybee}. 

Our approach offers three major advantages: 1) \textbf{Enhanced Visual Alignment}: By effectively modulating the context window length, we improve the attention weights allocated from loss tokens to vision tokens, leading to stronger visual grounding. 2) \textbf{Diverse Turn Distribution}: The method substantially diversifies both the distribution of turn counts and the historical content models see during training, enabling the model to handle various real-world scenarios with fluctuating context lengths. 3) \textbf{Efficient Data Augmentation}: The dynamic generation of turn variants serves as a powerful, compute-neutral data augmentation strategy that enhances model generalization while preserving training efficiency.
We validate our method across five distinct domain tasks~\cite{tan2025visionllmsbadhierarchical,han2025coralvqa,hsieh2025taiwanvqa,he2020pathvqa,liu2024mmdu} in both multiT and singleT evaluation settings. 

Finally, we note the performance gap between singleT and multiT evaluation exposes the brittle robustness of LVLMs, posing a risk when relying on either mechanism for LVLM assessment. Hence, we propose a novel evaluation mechanism, Balanced Latin Square Turn Permutation, that broadens the scope of evaluation beyond isolated singleT or multiT setting, focusing instead on the robustness of model outputs under various contextual perturbations.

We summarize our contribution as follows:
\begin{itemize}
\vspace{-10pt}
    \item To the best of our knowledge, we are the first to highlight and systematically analyze the  discrepancy between the common practice of multiT VIT and the prevalent singleT evaluation in the LVLM literature.
    \item  We propose StochasT, a compute-neutral data augmentation and training strategy that dynamically varies historical context to enhance LVLMs' visual alignment and contextual robustness.
    \item We propose Balanced Latin Square Turn Permutation, an evaluation mechanism to assess LVLMs' robustness under diverse contextual perturbations, moving beyond traditional singleT or multiT testing.
\end{itemize}

\section{Related Work}



\subsection{Visual Instruction Tuning}

Contemporary large vision-language models (LVLMs)~\cite{liu2023visual,zhu2023minigpt,dai2023instructblip,wang2024qwen2,bai2025qwen3vltechnicalreport,bai2025qwen25vltechnicalreport,zhu2025internvl3exploringadvancedtraining,wang2025internvl35advancingopensourcemultimodal,gemmateam2025gemma3technicalreport,grattafiori2024llama,team2025kimi,Liu_2024_CVPR} typically adopt the architecture that comprise a vision encoder~\cite{dosovitskiy2020image,radford2021learningtransferablevisualmodels,zhai2023sigmoidlosslanguageimage}, a cross-modal projector (e.g., an MLP or Q-Former~\cite{li2023blip}), and a pretrained large language model (LLM)~\cite{touvron2023llamaopenefficientfoundation,vicuna2023,qwen2025qwen25technicalreport,jiang2023mistral7b}. To elicit instruction-following capabilities over visual inputs, visual instruction tuning (VIT)~\cite{liu2023visual, zhu2023minigpt, dai2023instructblip, Liu_2024_CVPR} has emerged as the standard paradigm, analogous to language instruction tuning~\cite{wei2022finetunedlanguagemodelszeroshot,ouyang2022traininglanguagemodelsfollow,wang2023self}. VIT generally proceeds in two stages: (1) \textit{pretraining} the projector to align visual features with the text space, and (2) \textit{fine-tuning} the model via supervised fine-tuning (SFT), often followed by reinforcement learning (RL)~\cite{ouyang2022traininglanguagemodelsfollow,rafailov2023direct, shao2024deepseekmath} for enhanced alignment. Building upon this foundational pipeline, recent advancements primarily focus on refining training data and developing novel objectives. On the data front, studies demonstrate that rigorous curation is often more critical to VIT efficacy than sheer volume~\cite{zhou2023lima, wei2023instructiongpt4200instructionparadigmfinetuning}. Alongside these curation efforts, DRESS~\cite{chen2024dress} leverages natural language human feedback to enhance both alignment and multiT interaction capabilities. Beyond data-driven enhancements, several approaches modify training objectives to mitigate overfitting and improve generalization. L2T~\cite{zhou2025learning} introduces an auxiliary loss on instruction tokens to penalize shortcut learning and encourage the model to focus on visual tokens. Vittle~\cite{oh2025visual} formulates a novel learning objective grounded in the information bottleneck principle to increase robustness against distribution shifts. Furthermore, Ross~\cite{wang2025reconstructive} augments standard text supervision with a denoising objective that reconstructs latent visual representations, an approach recently generalized to 3D representation learning by Ross3D~\cite{wang2025ross3d}. Unlike prior works that neglect the balance between singleT and multiT capabilities during VIT, our approach naturally maintains an equilibrium without requiring auxiliary objectives or additional budget for data curation and training.


\subsection{Visual Instruction Tuning Datasets \& LVLM Evaluation}

VIT datasets generally consist of large-scale, general-purpose corpora (e.g., LLaVA-Instruct-665K~\cite{Liu_2024_CVPR}, MiniGPT-4~\cite{zhu2023minigpt}, SVIT~\cite{zhao2023svitscalingvisualinstruction}) and specialized downstream datasets. These specialized sets address targeted tasks such as document understanding~\cite{luo2024layoutllm}, hierarchical recognition~\cite{tan2025visionllmsbadhierarchical}, biomedical VQA~\cite{cui2024biomedical}, and various other domain-specific applications~\cite{han2025coralvqa,yuan2024osprey,pi2024personalized,hsieh2025taiwanvqa,Li_2025_CVPR,liu2024mitigating,lin2025comparison}. LVLMs instruction-tuned on these data are predominantly evaluated in singleT settings using standard benchmarks like MMMU~\cite{yue2024mmmu}, MathVista~\cite{lu2023mathvista}, MMBench~\cite{liu2024mmbench}, MMStar~\cite{chen2024we}, and MME~\cite{fu2025mme}. Concurrently, recent research has increasingly focused on eliciting the multiT conversational capabilities of LVLMs~\cite{liu2024mmdu,yan2025mmcr,feng2023mmdialog,lei2025contextqformer,chen2025intermt}, prompting the development of benchmarks specifically tailored for multiT interactions~\cite{liu2024mmdu,liu2024convbench,liu-etal-2025-taking,1013332}. Despite these advancements, a critical gap remains: the absence of training recipes and balanced evaluation metrics designed to jointly foster and assess both singleT and multiT paradigms. 

\section{Method}
\subsection{SingleT vs.\ MultiT Visual Instruction Tuning (VIT)}
\label{sec:problem_formulation}

A Large Vision-Language Model (LVLM), parameterized by $\theta$, is designed to process a visual input $X_v$ and a text instruction $X_q$, generating a corresponding textual response $X_a$ of length $L$. This generation process is modeled autoregressively as:
\begin{equation}
    P_\theta(X_a | X_v, X_q) = \prod_{i=1}^{L} P_\theta(x_{a,i} | X_v, X_q, x_{a,<i}),
\end{equation}
where $x_{a,i}$ is the $i$-th token of the response $X_a$, and $x_{a,<i}$ denotes the sequence of preceding tokens.

In the context of multiT VIT, the model processes a conversational dialogue $\mathcal{C} = \{(X_q^{(n)}, X_a^{(n)})\}_{n=1}^N$ based on the image $X_v$, where $N$ denotes the total number of dialogue turns. At the $n$-th turn, $X_q^{(n)}$ represents the user instruction and $X_a^{(n)}$ represents the target response. Under the standard multiT paradigm, the sequence is packed, and the objective is to minimize the negative log-likelihood of the response tokens across all $N$ turns:
\begin{equation}
    \mathcal{L}_{\text{multi}} = -\hlv{{\sum_{n=1}^{N}}} \sum_{i=1}^{L_n} \log P_\theta(x_{a,i}^{(n)} | X_v, \hlv{H^{(n)}}, x_{a,<i}^{(n)}),
\end{equation}
where $L_n$ is the sequence length of the $n$-th response, and $H^{(n)}$ denotes the accumulated conversational context prior to the $n$-th response, defined as $H^{(n)} = [X_q^{(1)}, X_a^{(1)}, \dots, X_q^{(n)}]$. Pioneering frameworks, such as LLaVA~\cite{liu2023visual}, synthesize these $N$-turn dialogues $\mathcal{C}$ using advanced large language models. The generated instructions $X_q^{(n)}$ encompass a diverse array of visual tasks, ranging from object categorization and counting to spatial reasoning and action recognition. By optimizing $\mathcal{L}_{\text{multi}}$, the LVLM learns instruction-following capabilities across diverse contexts within a single forward pass.

However, because the $N$ turns within the dialogue $\mathcal{C}$ often focus on disparate visual features with minimal semantic dependency, an alternative approach is to unroll the multiT dialogue into $N$ independent singleT samples, defined as $\mathcal{S} = \{(X_v, X_q^{(n)}, X_a^{(n)})\}_{n=1}^N$. In this singleT paradigm, the historical context is discarded such that $H^{(n)} = X_q^{(n)}$, and the model optimizes an unrolled objective over the independent pairs:
\begin{equation}
    \mathcal{L}_{\text{single}} = -\hlv{\sum_{n=1}^{N}} \sum_{i=1}^{L_n} \log P_\theta(x_{a,i}^{(n)} | X_v, \hlv{X_q^{(n)}}, x_{a,<i}^{(n)}).
\end{equation}
While structurally simpler, duplicating $X_v$ across $N$ isolated samples is computationally less efficient than the packed multiT sequence. Despite this inefficiency, the singleT setting is still adopted in the training setups of some prior works~\cite{chen2024dress,cha2024honeybee} and in the organization of several datasets~\cite{liu2024mitigating,han2025coralvqa}. 

This structural divergence raises a critical, largely overlooked research question: \textit{How do the multiT ($\mathcal{L}_{\text{multi}}$) and singleT ($\mathcal{L}_{\text{single}}$) training paradigms differentially impact an LVLM's intrinsic capabilities and its  performance on downstream tasks?} {In this paper, we systematically investigate this dynamic and propose a robust visual instruction tuning strategy that harmonizes both paradigms, alongside a novel evaluation strategy utilizing two new metrics to benchmark model robustness.}

\subsection{Stochastic Turn Depth}
\label{sec:scd}
\begin{figure}[t]
    \centering
    \includegraphics[width=1\linewidth]{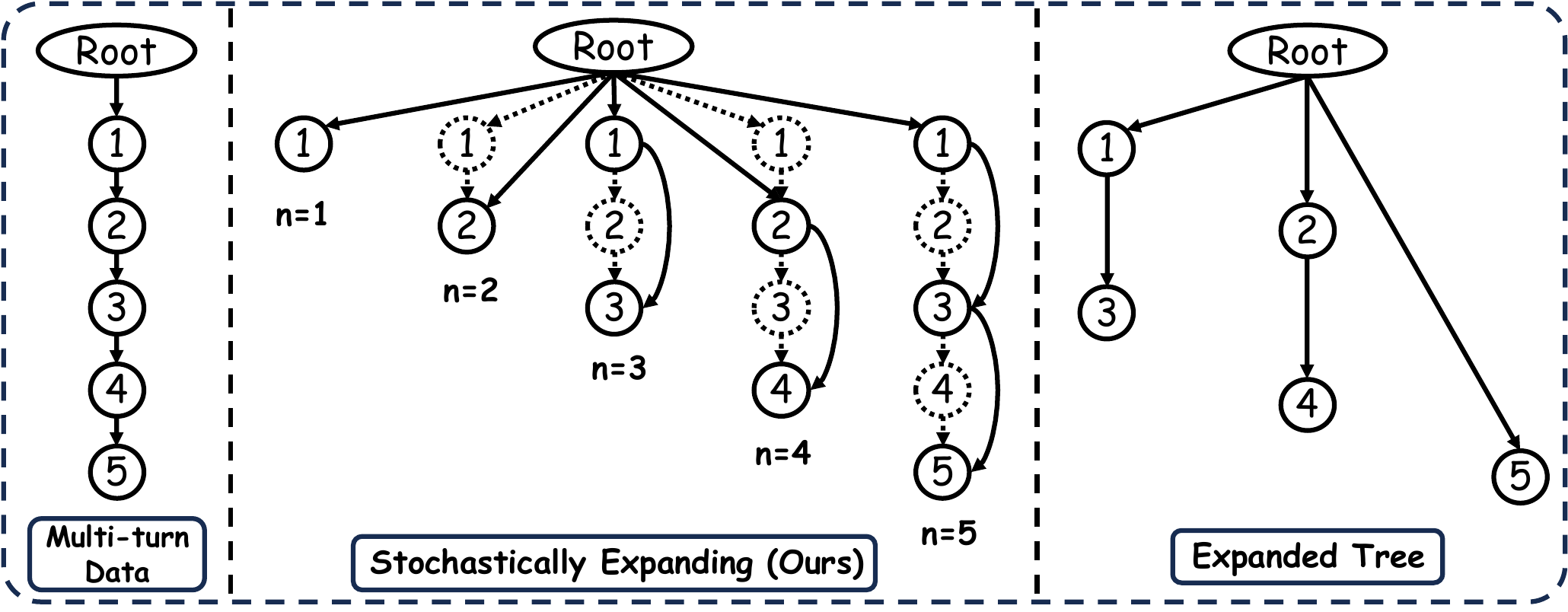}
    \caption{\textbf{An overview of our StochasT procedure.} Starting from the first turn of a multiT dialogue, the method randomly omits previous context turns to generate an expanded conversation tree for the training process. The root contains the system prompt and the image grounding the turns.}
    
    \label{fig:method}
\end{figure}

Dropout~\cite{srivastava2014dropout} is a fundamental regularization technique widely adopted in deep neural networks. It randomly deactivates neurons during training, implicitly creating an ensemble of networks with varying capacities. Beyond individual neurons, this principle has been extended to entire architectural blocks. For instance, Stochastic Depth~\cite{huang2016deep} randomly drops residual blocks within deep ResNets to implicitly train an ensemble of networks with varying depths, demonstrating superior performance over standard training paradigms.
Building on this idea, we propose Stochastic Turn Depth (StochasT). Specifically, we perform visual instruction tuning by stochastically pruning the historical context of each turn during training. Learning with stochastic turn depth can therefore be interpreted as training an LVLM over an implicit ensemble of conversational trajectories with dynamically varying context lengths. Under this unified framework, the standard singleT and multiT training paradigms naturally arise as boundary cases.

For a training dataset $\mathcal{D}$ comprising multiT dialogues in the form of $\mathcal{C} = \{(X_q^{(n)}, X_a^{(n)})\}_{n=1}^N$, the most intuitive approach to implementing StochasT is to stochastically drop individual turns from each conversation. 
We assign a retention probability to each turn, analogous to standard dropout mechanisms. Specifically, for the $n$-th turn, we define a drop probability $p_n$ sampled from a Beta distribution:
\begin{equation}
    p_n \sim \text{Beta}(\alpha, \beta),
\end{equation}
where $\alpha$ and $\beta$ are hyperparameters controlling the shape of the distribution, allowing us to flexibly bias the dropout rate toward specific interaction lengths. We then sample a binary indicator mask $m_n \sim \text{Bernoulli}(1 - p_n)$ for each turn $n \in \{1, \dots, N\}$. 
Therefore, after sampling, the original training dialogue $\mathcal{C}$ is reduced to a stochastically pruned conversation $\hat{\mathcal{C}}$, formally defined as:
\begin{equation}
\label{beta}
    \hat{\mathcal{C}} = \left\{ (X_q^{(n)}, X_a^{(n)}) \mid m_n = 1, \text{ for } n=1, \dots, N \right\}.
\end{equation}
In this formulation, if $m_n = 0$, the $n$-th instruction-response pair is entirely removed from the sequence for the current training iteration. Consequently, the historical context $H$ for any subsequent retained turn $k > n$ is dynamically altered. 


\setlength{\intextsep}{6pt}   
\setlength{\columnsep}{8pt}   
\begin{wrapfigure}{r}{0.48\textwidth}
\begin{minipage}{0.48\textwidth}
\begin{algorithm}[H]
\caption{Stochastic Turn Depth}
\label{alg:tree_expansion}
\textbf{Input}: $N$ dialogue turns, Beta parameters $\alpha, \beta$.\\
\textbf{Output}: Parent node assignments $P$.
\begin{algorithmic}[1]
\STATE [\textbf{Optional}] Randomly shuffle turns
\FOR{$n = 1$ \TO $N$}
    \STATE $k \leftarrow n - 1$
    \WHILE{$k > 0$}
        \STATE $p_k \sim \text{Beta}(\alpha, \beta)$, 
        \STATE $m_k \sim \text{Bernoulli}(1 - p_k)$
        \IF{$m_k$} \STATE \textbf{break} \ENDIF
        \STATE $k \leftarrow k - 1$
    \ENDWHILE
    \STATE $P[n] \leftarrow k$ \COMMENT{$k=0$ connects to root $X_v$}
\ENDFOR
\STATE \textbf{return} $P$
\end{algorithmic}
\end{algorithm}
\end{minipage}
\end{wrapfigure}

While directly dropping  simplifies data preprocessing and accelerates training by physically reducing the sequence length, it inherently decreases the number of effective tokens available for loss calculation per training sample. To maximize token utilization while maintaining context diversity, we propose {learning with stochastic depth (StochasT)}. Instead of explicitly removing turns from the input sequence, we retain all $N$ turns of the dialogue $\mathcal{C}$ for loss computation but stochastically mask their historical dependencies. 

Due to the inherent autoregressive architecture of LVLMs, sequence interactions are governed by a causal attention mask and positional embeddings, which grant each token a unique positional index. We leverage this architecture to perform turn dropout in the backward direction, thereby preserving strict chronological causality. For each turn $n \in \{1, \dots, N\}$, rather than assigning a deterministic, sequential history $H^{(n)}$, we stochastically connect the turn to a sampled parent node $P(n)$ to construct a diversified history $\tilde{H}^{(n)}$. Consequently, this procedure can be formulated as a directed graph expansion. We incrementally build this dialogue tree from a root node representing all tokens prior to the first turn. By convention, we designate the initial visual input $X_v$ as this root.

Formally, to determine the parent $P(n)$ for the $n$-th turn, we traverse backward through its preceding nodes $k = n-1, n-2, \dots, 1$. At each step $k$, we sample a drop probability $p_k \sim \text{Beta}(\alpha, \beta)$ and an indicator variable $m_{n,k} \sim \text{Bernoulli}(1 - p_k)$. The backward traversal halts at the first node where $m_{n,k} = 1$, establishing it as the parent node to which turn $n$ connects. If all preceding nodes are dropped (i.e., $m_{n,k} = 0$ for all $k < n$), the turn connects directly to the root, effectively setting $P(n) = 0$. The resulting procedural logic for constructing this stochastic tree is detailed in \cref{alg:tree_expansion}.

By executing this algorithm, the sequentially packed multiT dialogue expands into a structured tree graph, as illustrated in \cref{fig:method}. {Our method} effectively resolves the trade-off between the singleT and multiT paradigms. Compared to the flat structure of singleT training $\mathcal{S}$, our method provides greater depth and complex contextual dependencies (preserving multiple nodes along specific branches). Conversely, compared to the rigid chain of standard multiT training $\mathcal{C}$, it introduces greater breadth and contextual diversity while strictly respecting the chronological causality of the original dialogue. Note that when the drop probability $p_n \to 1$ for all turns, StochasT naturally degenerates into the singleT paradigm $\mathcal{S}$, whereas setting $p_n = 0$ strictly recovers the standard multiT paradigm $\mathcal{C}$. Ultimately, this approach reaches a robust equilibrium, optimizing the LVLM's capabilities across varying conversational lengths. {Note that StochasT is best suited for the conversations where turns are not strongly interdependent.}

It is instructive to contrast our StochasT tree expansion with the well-known Chinese Restaurant Process (CRP)~\cite{teh2004sharing}. While both are discrete stochastic processes that incrementally assign sequential entities (turns versus customers) to previous states to form complex structural dependencies, their underlying mechanisms and objectives diverge significantly. The CRP models cluster assignment through preferential attachment. Each new customer selects a table with probability proportional to its current occupancy, without considering the temporal order of previous arrivals. This naturally yields un-ordered, disjoint partitions. In contrast, StochasT constructs a causal directed tree by enforcing a strict chronological backward traversal. Rather than relying on cluster popularity, our stochastic assignment is governed by consecutive Bernoulli trials tied to temporal proximity. This critical distinction ensures that the autoregressive causality and sequence dependency inherent to LVLMs are rigorously preserved while still injecting the desired structural stochasticity.


\section{Balanced Latin Square Turn Permutations}
\label{sec:metric}

A general-purpose LVLM should yield consistently correct responses to an instruction regardless of whether it is posed in isolation or embedded within a complex, multiT dialogue. However, \cref{fig:stat} reveals a significant performance gap between singleT and multiT evaluation, indicating that models are highly sensitive to domain shifts and contextual variations. Standard evaluation metrics, which compute raw accuracy on static question-answer pairs, fail to capture this vulnerability. While some recent works~\cite{duan2024vlmevalkit,hsieh2025taiwanvqa} employ circular (of choices) evaluation on multi-choice questions, directly extending it from choices to turns brings in limited contextual variation; with the original inter-turn adjacency intact.

To rigorously assess model robustness under varying contextual dependencies, we propose a novel evaluation paradigm based on the Balanced Latin Square (BLS)~\cite{f9811b4c-b3c3-3066-a566-620f6d88748c}. BLS constructs an $N \times N$ matrix where each turn appears exactly once in every dialogue position, and every turn immediately precedes every other turn exactly once. Hence, evaluating LVLMs following these permutations can systematically control for both absolute positional bias and first-order carry-over effects.
A standard BLS requires $N$ to be an even number; if $N$ is odd, fully balancing the permutations requires a $2N \times N$ matrix, which doubles the inference budget. To maintain computational efficiency, if the number of evaluation turns $N$ is odd, we pad the dialogue sequence with a universal placeholder instruction: $X_q^{(\text{dummy})} = \texttt{Please briefly describe this image.}$ This placeholder does not contribute to final accuray. This ensures the effective sequence length $\tilde{N}$ is always even (where $\tilde{N} = N$ if $N$ is even, and $\tilde{N} = N + 1$ if $N$ is odd), allowing us to construct a perfectly balanced $\tilde{N} \times \tilde{N}$ permutation matrix corresponding to exactly $\tilde{N}$ inferences. 

Let $c_{n,j} \in \{0, 1\}$ denote the binary correctness of the model's response to the $n$-th turn when evaluated under the $j$-th permutation sequence. We introduce two novel metrics to quantify a model's intrinsic capabilities.

\textbf{Context-Robust Accuracy (CRA).}
This metric computes the average accuracy of a specific turn across all $\tilde{N}$ contextual permutations,
\begin{equation}
    \text{CRA}^{(n)} = \frac{1}{\tilde{N}} \sum_{j=1}^{\tilde{N}} c_{n,j} \quad \text{and} \quad \text{CRA}=\frac{1}{\tilde{N}}\sum_{n=1}^{\tilde{N}} \text{CRA}^{(n)},
\end{equation}
which measures a model's robustness under distinct contextual histories.

\textbf{Strict Context-Robust Accuracy (CRA+).} We then define a more stringent metric, 
\begin{equation}
    \text{CRA+}^{(n)} = \prod_{j=1}^{\tilde{N}} c_{n,j} \quad \text{CRA+} \;=\;\frac{1}{\tilde{N}}\sum_{n=1}^{\tilde{N}} \text{CRA+}^{(n)},
\end{equation}
where $\text{CRA+}^{(n)}$ is 1 only when the model makes no mistake under all the turn histories / permutations. 
It assess the model's genuine, context-invariant knowledge regarding the visual instruction.

\medskip
\noindent\textbf{Limitation and Discussion.}
The proposed evaluation mechanism assumes the multiple instruction-answer turns about an image are independent of each other, to some extent. It holds for a majority of existing LVLM evaluation benchmarks, to the best of our knowledge. Still, it is important to note that one should not apply it to conversations with strong dependency on historic contexts. 



\section{Experiments}
\label{sec:experiments}

We validate our proposed StochasT framework using Qwen2.5-VL-3B~\cite{bai2025qwen25vltechnicalreport} and LLaVA-1.5-7B~\cite{Liu_2024_CVPR}. Following the experimental setup detailed in \cref{subsec:setup}, we comprehensively compare our approach against standard  multiT and singleT training baselines across a diverse suite of downstream tasks and leverage we our novel Balanced Latin Square metrics to assess models' robustness (\cref{subsec:main_results}). In~\cref{subsec:ablation_analysis}, we conduct ablation studies on the dropout distribution hyperparameters, and systematically investigate the impact of StochasT on visual attention.

\subsection{Experimental Setup}
\label{subsec:setup}

\textbf{Datasets.} We select diverse VIT datasets spanning multiple domains, specifically targeting downstream tasks that inherently feature multiT instructions per image. \textbf{iNat-Plant}~\cite{tan2025visionllmsbadhierarchical} is a hierarchically constructed instruction tuning dataset designed for fine-grained visual understanding, grounded in the plant taxonomy of iNaturalist-2021~\cite{van2021benchmarking}. \textbf{PathVQA}~\cite{he2020pathvqa} is a pioneering dataset focused on clinical pathology. As it lacks an explicit visual instruction tuning subset, we programmatically reformat its original training split into a standard instruction-response format. \textbf{CoralVQA}~\cite{han2025coralvqa} is curated to develop and assess the specialized capability of LVLMs in coral reef analysis. \textbf{TaiwanVQA}~\cite{hsieh2025taiwanvqa} targets the adaptation of LVLMs to culturally specific visual content, evaluating both localized recognition and nuanced reasoning. Additionally, we validate our approach on \textbf{MMDU}~\cite{liu2024mmdu}, an explicit multiT, multi-image instruction tuning dataset designed to enhance an LVLM's extended conversational capabilities and complex multi-image reasoning.

\medskip\noindent\textbf{Benchmarks and Evaluation.} We evaluate the trained models on their respective official benchmarks or test splits. For the multiple-choice tasks in iNat-Plant and TaiwanVQA, we directly extract the model's predicted options. For the open-ended generative tasks in MMDU, CoralVQA, and PathVQA, we employ Gemini 3 Flash to evaluate the generated responses. 

\medskip\noindent\textbf{Implementation Details.} We apply LoRA~\cite{hu2022lora} fine-tuning to the official visual instruction-tuned checkpoints of Qwen2.5-VL and LLaVA-1.5. Our training pipeline leverages the official LLaVA-1.5 codebase and the designated fine-tuning repository~\cite{Qwen2-VL-Finetuning} for Qwen2.5-VL. Comprehensive hyperparameters, hardware configurations, and training specifics are detailed in the supplement.

\subsection{Main Results}
\label{subsec:main_results}

\begin{table}[t]
\centering
\caption{Performance comparison of different training strategies on LLaVA-1.5-7B and Qwen2.5-VL-3B under singleT and multiT evaluation settings.}
\label{tab:model_comparison_compact}
\renewcommand{\arraystretch}{1.15} 
\setlength{\tabcolsep}{0.7pt}        
\begin{tabular}{@{}ll | l  cc@{}}
\toprule
\textbf{Model} & \textbf{Eval.} & \textbf{Training} & \textbf{Avg. Accuracy over~\cite{he2020pathvqa,han2025coralvqa,tan2025visionllmsbadhierarchical,hsieh2025taiwanvqa}} & \textbf{MMDU~\cite{liu2024mmdu}} \\
\midrule

\multirow{8}{*}{LLaVA-1.5-7B} 

& \multirow{4}{*}{SingleT} 
  & Original           & 43.74 & -- \\
& & MultiT                  & 61.51 & -- \\
& & SingleT                 & \textbf{68.46} & -- \\
& & \textbf{StochasT}& 67.16 & -- \\
\cmidrule(l){2-5}
& \multirow{4}{*}{MultiT}  
  & Original          & 46.84 & 10.81 \\
& & MultiT                  & 68.52 & 12.68 \\
& & SingleT                 & 68.60 & --   \\
& & \textbf{StochasT}& \textbf{70.63} & \textbf{13.10} \\

\midrule

\multirow{8}{*}{Qwen-2.5-VL-3B}
& \multirow{4}{*}{SingleT} 
  & Original           & 55.48 & -- \\
& & MultiT                  & 71.20 & -- \\
& & SingleT                 & 76.71 & -- \\
& & \textbf{StochasT}& \textbf{77.28} & -- \\
\cmidrule(l){2-5}
& \multirow{4}{*}{MultiT}  
  & Original           & 58.16 & 47.00 \\ 
& & MultiT                  & 77.39 & 57.90 \\ 
& & SingleT                 & 77.75 & --   \\
& & \textbf{StochasT}& \textbf{80.16} & \textbf{59.80} \\

\bottomrule
\end{tabular}

\end{table}

We compare our method against the standard multiT visual instruction tuning baseline  as well as a specialized singleT training setting. The evaluations are conducted under both singleT and multiT settings, as detailed in \cref{tab:model_comparison_compact}. We report the average accuracy across all four datasets, while listing MMDU separately as it does not support singleT training and evaluation. Detailed results for each individual dataset {and Qwen3-VL-32B~\cite{bai2025qwen3vltechnicalreport}} are provided in the supplement. 

As shown in \cref{tab:model_comparison_compact}, our method consistently outperforms MultiT across all evaluation settings for both models. Specifically, StochasT yields an average performance gain of 9.19$\%$ relative to singleT evaluation and 3.08$\%$ over multiT evaluations for LLaVA-1.5-7B, alongside gains of 8.54$\%$ and 3.58$\%$, respectively, for Qwen2.5-VL-3B. As expected, SingleT also outperforms MultiT across all singleT evaluations due to the in-distribution nature of this testing. Additionally, SingleT achieves comparable performance to MultiT on multiT evaluations. We conjecture that this is caused by the effectively shorter sequence lengths in context, allowing loss tokens to attend more efficiently to the visual inputs and instructions. We discuss this dynamic in detail in the subsequent section.

\begin{figure}[t]
    \centering
    \includegraphics[width=1\linewidth]{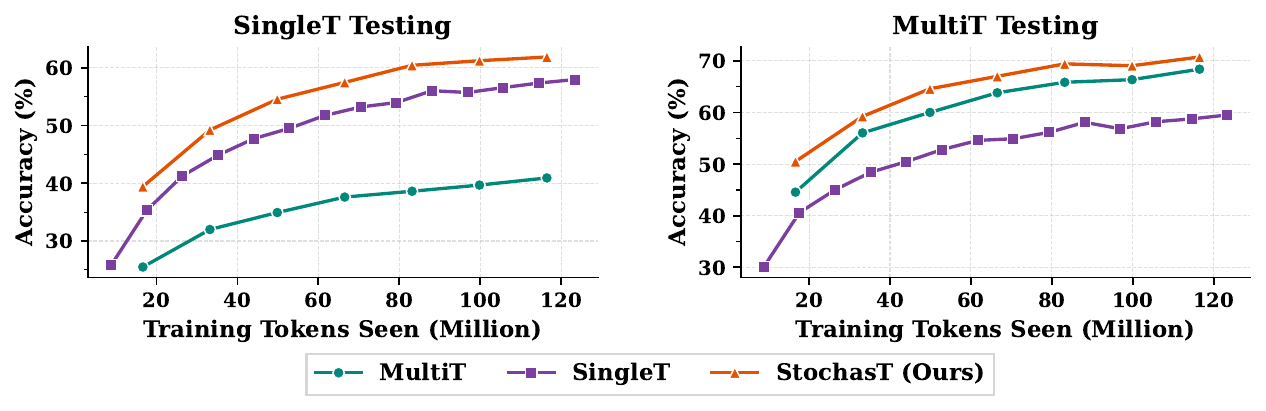}
    \caption{\textbf{Accuracy vs. \#training tokens for different training strategies.} We evaluate MultiT, SingleT, and our proposed StochasT under singleT and multiT testing. As the number of training tokens increases, StochasT consistently outperforms both baselines and shows faster convergence, especially in the low-token regime. }
    \label{fig:tokens}
\end{figure}

Compared to SingleT, our method achieves state-of-the-art (SOTA) performance on all multiT evaluations across both models, while remaining highly competitive on singleT evaluations. Importantly, as illustrated in \cref{fig:tokens}, SingleT requires approximately double the number of tokens to achieve similar singleT performance. This inefficiency arises from data unrolling, where the same image is duplicated multiple times within a single training epoch, leading to a massive influx of redundant tokens. This contrast highlights the perfect equilibrium StochasT achieves between training efficiency and downstream performance.

\medskip\noindent\textbf{SingleT vs.\ MultiT Evaluation.} 
The MultiT baseline exhibits a noticeable vulnerability when evaluated on isolated singleT instructions, struggling to generalize outside of a continuous conversational context. multiT evaluation consistently outperforms singleT evaluation, exposing an average gap of 7.01$\%$ on LLaVA-1.5-7B and 6.19$\%$ on Qwen2.5-VL-3B. Notably, the original gap before downstream instruction tuning was less than half of this magnitude (3.10$\%$ and 2.68$\%$, respectively). This indicates that standard multiT training inherently exacerbates this discrepancy, rendering the models increasingly brittle to contextual perturbations. With StochasT, this performance gap narrows significantly to 3.47$\%$ and 3.33$\%$, closely resembling the baseline robustness of the pre-trained models. This demonstrates the effectiveness of our approach in maintaining original structural robustness while heavily regularizing the learned distribution. For SingleT, this gap is reduced to less than 1$\%$, making it an ideal debiasing choice exclusively in scenarios where computational overhead is not a concern.

\setlength{\intextsep}{4pt}   
\setlength{\columnsep}{8pt}   
\begin{wraptable}{l}{0.48\textwidth}
\centering
\caption{The context-robust accuracy (CRA) and strict context-robust accuracy (CRA+) averaged over four datasets.}
\footnotesize
\label{tab:BLS}
\begin{tabular}{lcc}
\toprule
\textbf{Method} & \textbf{CRA} & \textbf{CRA+} \\
\midrule

LLaVA-1.5-7B & 45.73 & 27.66 \\
\quad +MultiT & 61.82 & 41.19 \\
\quad +SingleT & 67.33 & \textbf{53.15} \\
\quad +\textbf{StochasT (Ours)} & \textbf{67.69} & 50.89 \\

\midrule

Qwen2.5-VL-3B & 57.26 & 38.46 \\
\quad +MultiT & 73.63 & 54.30 \\
\quad +SingleT & 77.26 & \textbf{64.57} \\
\quad +\textbf{StochasT (Ours)} & \textbf{77.53} & 61.76 \\

\bottomrule
\end{tabular}
\end{wraptable}
\medskip\noindent\textbf{MultiT Multi-Image Training.} 
Surprisingly, StochasT even outperforms MultiT on MMDU, a complex long-context dialogue benchmark with related sequential turns. {We conjecture that this advantage stems from the relatively weak interdependency among MMDU turns, as observed in~\cite{yan2025mmcr}. Under such mild dependencies, StochasT remains robust by preserving the original turn order while encouraging stronger visual grounding rather than relying on brittle textual shortcuts.}


\medskip\noindent\textbf{Balanced Latin Square Evaluation.} 
\Cref{tab:BLS} presents the results of our Balanced Latin Square evaluation. On both models, StochasT achieves the highest Context-Robust Accuracy (CRA), indicating comprehensively superior performance across diverse historical contexts. This demonstrates its immense utility as a foundational training strategy for real-world deployment, and we advocate for CRA as a primary metric for generalized LVLM evaluation. For Strict Context-Robust Accuracy (CRA+), which requires consistency under all contextual perturbations, SingleT marginally outperforms StochasT, since training with repeatedly duplicated images forces the model toward highly deterministic outputs through brute-force reinforcement. Nevertheless, this establishes CRA+ as an excellent proxy for an LVLM's predictive uncertainty.


\subsection{Ablation and Analysis}
\label{subsec:ablation_analysis}

\medskip\noindent\textbf{{Ablation on Dropout Distribution.}}
We adopt the Beta distribution to sample dropout rates because it flexibly models bounded probabilities with only two parameters.
By varying $(\alpha,\beta)$, it can smoothly interpolate between SingleT-like shallow histories and MultiT-like deep histories, making it a simple and practical default among possible bounded distributions. We compare different $(\alpha,\beta)$ settings that emphasize varying amounts of contextual history: higher dropout rates reduce history and encourage visual grounding, while lower dropout rates preserve deeper conversations for long-context reasoning. By default, all main experiments use a symmetric unimodal distribution with $\alpha=2$ and $\beta=2$.

We evaluate four distinct configurations on CoralVQA and iNat-Plant: a U-shaped distribution $(0.5,0.5)$, a lower-biased distribution $(1,5)$, a higher-biased distribution $(5,1)$, and our default symmetric setting $(2,2)$. 

\setlength{\intextsep}{4pt}   
\setlength{\columnsep}{8pt}   
\begin{wraptable}{r}{0.65\textwidth}
    \centering
    \small
        \caption{\textbf{Left}: Effect of Beta parameters $(\alpha,\beta)$ on model performance. \textbf{Right}: Comparison with random-fixed cutoff (RC) baseline, averaged over two datasets.}
    \label{tab:ablation}
    \setlength{\tabcolsep}{1pt}
    \begin{tabular}{l | cccc | ccc}
    \toprule
    \multirow{2}{*}{Metric} 
    & \multicolumn{4}{c|}{Beta Parameters $(\alpha,\beta)$} 
    & \multicolumn{3}{c}{Qwen2.5VL-3B} \\
    \cmidrule(lr){2-5} \cmidrule(lr){6-8}
    & (0.5,0.5) & (2,2) & (1,5) & (5,1) 
    & MultiT & RC & \textbf{Ours} \\
    \midrule
    CRA  & 80.92 & \textbf{82.26} & 81.04 & 81.98 
         & 73.65 & 70.61 & \textbf{79.24} \\
    CRA+ & 69.05 & \textbf{73.30} & 69.81 & 72.69 
         & 44.05 & 44.95 & \textbf{56.41} \\
    \bottomrule
    \end{tabular}
\end{wraptable}
\noindent As shown in \cref{tab:ablation} (Left), performance remains relatively robust across the parameter space,with the symmetric $(2,2)$ setting achieving the best overall results. Notably, $(5,1)$, which favors shorter SingleT-like histories, consistently outperforms $(1,5)$, which is closer to MultiT. {Overly suppressing dropout makes the framework regress toward deterministic MultiT training, reintroducing contextual overfitting and reducing the gains under the stringent CRA+ metric.}

\medskip\noindent\textbf{{Ablation on Turn Depth.}}
{To examine whether the gains simply arise from a smaller expected turn depth, we compare StochasT with a random-fixed cutoff (RC) baseline on iNat-Plant and PathVQA, matching their expected turn depths and reporting the averaged results in~\cref{tab:ablation} (Right). Directly shortening the turn depth underperforms even MultiT, suggesting that StochasT's improvements do not merely result from a shorter effective context or generic context perturbation.}

\medskip\noindent\textbf{Impact of StochasT on Visual Attention.}
In LVLMs, visual tokens typically occupy a vast portion of the attention matrix while receiving a disproportionately small fraction of the total attention weights~\cite{chen2024image,zhou2025learning,an2025mitigating}, whereas text tokens dominate the attention distribution. This imbalance is a widely observed bottleneck. Moreover, as the sequence length increases, this effect becomes increasingly severe, formalized by Yang~\etal~\cite{yang2025ikod} as visual attention degradation.

To investigate whether StochasT effectively forces the model to better utilize visual information, we evaluate our approach using the Visual Contribution (VC) metric proposed by Zhou~\etal~\cite{zhou2025learning}. VC quantifies visual reliance by measuring the difference in output negative log-likelihood between a standard forward pass and a perturbed forward pass where the image is replaced by random noise. We compute VC over 1,000 samples each from the iNat-Plant and GQA validation splits using LLaVA-1.5-7B trained on iNat-Plant.

\begin{figure}[t]
    \centering
    \includegraphics[width=1\linewidth]{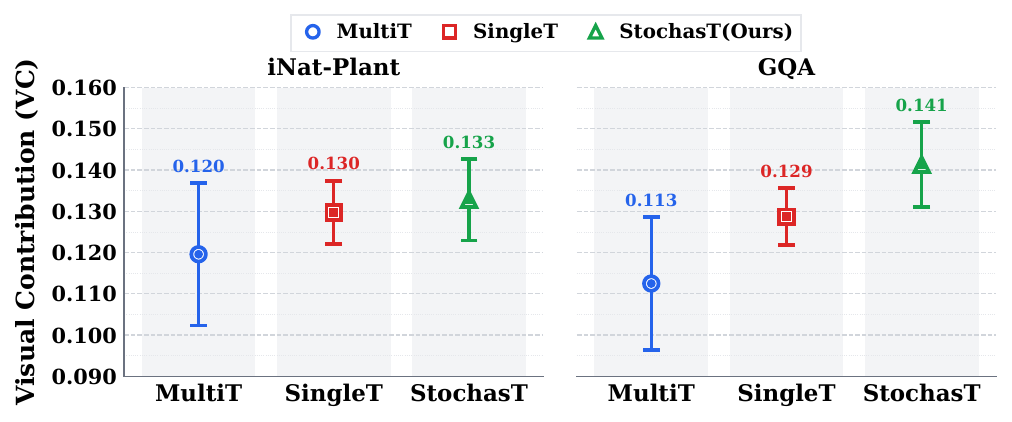}
    \caption{\textbf{Visual contribution (VC) on GQA and iNat-Plant.} Mean VC and variance for MultiT, SingleT, and StochasT on iNat and GQA. Our method consistently increases visual contribution, suggesting improved visual grounding.}
    \label{fig:vc}
\end{figure}

As illustrated in \cref{fig:vc}, StochasT achieves the highest mean VC on both datasets for both task-specific and general instructions. This indicates that StochasT successfully conditions the model to anchor its generation more heavily on visual features. Additionally, MultiT exhibits the lowest mean and the highest standard deviation, rendering it highly vulnerable to input perturbations and elevating its risk of hallucination. SingleT maintains a moderate VC with a small standard deviation, indicating strong robustness to input perturbations, which firmly aligns with its high CRA+ score identified earlier.

\section{Conclusion}
\label{sec:conclusion}

In this work, we highlight and systematically analyze a critical, previously overlooked vulnerability in Large Vision-Language Models (LVLMs): the severe performance discrepancy arising from the structural mismatch between multiT training and singleT evaluation. To bridge this gap, we introduce StochasT, an elegant and efficient training paradigm that implicitly optimizes the model over an ensemble of dynamically varying context lengths via the stochastic dropping of conversational history. Our extensive evaluations demonstrate that StochasT successfully harmonizes the strengths of both single- and multiT training. By penalizing reliance on spurious textual shortcuts, our method compels the LVLM to anchor its reasoning directly on the underlying visual features, thereby significantly enhancing multimodal utilization. Furthermore, we propose the Balanced Latin Square Turn Permutation evaluation framework, accompanied by two metrics: Context-Robust Accuracy (CRA) and Strict Context-Robust Accuracy (CRA+). Ultimately, these contributions not only expose the inherent brittleness of existing LVLMs but also establish a comprehensive, rigorous framework for developing inherently robust and context-resilient multimodal architectures.


\section*{Acknowledgements}
This work was supported in part by NSF 2540851 and a Gemini Academic Program Award. 

%
%
\bibliographystyle{splncs04}
\bibliography{main}

\clearpage
\appendix
\section*{\centering \Large Supplementary Material}
\renewcommand\thesection{\Alph{section}}
\renewcommand\thesubsection{\thesection.\arabic{subsection}}

\noindent This supplement provides comprehensive technical details, additional empirical evaluations, and qualitative analyses to support the findings of the main paper. In~\cref{sec:eval_protocol}, we elaborate on our formal evaluation protocols, detailing the exact setups for the single-turn, multi-turn, and our proposed Balanced Latin Square (BLS) evaluation frameworks. \Cref{sec:implementation} outlines the comprehensive implementation details, including dataset statistics and training configurations. In~\cref{sec:additional_exp}, we present extended experimental results to further validate our approach; this includes a direct comparison against a vanilla Turn Dropout baseline, and an additional study on extending StochasT to the gneral visual instruction tuning stage on the LLaVA-150K dataset in~\cref{sec:full_finetuning}. Finally, \Cref{sec:qualitative} provides qualitative examples that visually demonstrate the enhanced contextual robustness and context-invariant stability achieved by our method compared to standard baselines.

\section{Evaluation Protocol}
\label{sec:eval_protocol}

\subsection{Single-Turn Evaluation Setup}
For this evaluation, we follow the standard protocol of processing each question related to the same image independently in separate forward passes. The final single-turn (SingleT) performance is calculated by averaging the accuracy across all individual turns.

\subsection{Multi-Turn Evaluation Setup}
For the multi-turn (MultiT) setup, we evaluate each question within the dataset sequentially. The historical context for every step is constructed by concatenating all preceding questions and their corresponding \textbf{model answers}. Consistent with the SingleT protocol, the final performance metric is calculated by averaging the accuracy across all individual turns. However, this sequential evaluation is highly sensitive to the specific ordering of the questions. Given that the semantic connections between consecutive turns are typically weak in standard visual instruction tuning datasets, this sensitivity introduces significant evaluation noise. This inherent vulnerability directly motivates the introduction of our Balanced Latin Square (BLS) evaluation framework.

\subsection{Balanced Latin Square (BLS) Evaluation Setup}
An example of a $4 \times 4$ Balanced Latin Square is shown in \cref{equa:bls}. If we denote the indices of a four-turn conversation as $\{1, 2, 3, 4\}$, each row in the matrix represents the specific order in which the turns are evaluated. Consequently, a four-turn conversation requires four separate multi-turn evaluation passes. 

Notably, the first row represents the original MultiT evaluation order. Conversely, the first column demonstrates that each of the four turns appears exactly once in the starting position (with no prior history), which perfectly replicates the SingleT setup. By structuring the evaluation this way, we comprehensively encompass both SingleT and MultiT paradigms while systematically eliminating first-order carry-over effects. This drastically reduces evaluation noise and isolates the model's true contextual robustness.

\begin{equation}
\label{equa:bls}
\renewcommand{\arraystretch}{1.5} 
\setlength{\arraycolsep}{10pt}    
L = \begin{bmatrix}
\mathbf{1} & \mathbf{2} & \mathbf{3} & \mathbf{4} \\
\mathbf{2} & 4 & 1 & 3 \\
\mathbf{3} & 1 & 4 & 2 \\
\mathbf{4} & 3 & 2 & 1
\end{bmatrix}
\end{equation}
\subsection{Sanity Check on Dummy-Prompt Padding}
To verify that odd-$N$ padding does not confound CRA/CRA+, we evaluate naturally even-$N$ subsets of CoralVQA and PathVQA without padding, and then inject two dummy prompts for re-evaluation. The averaged CRA/CRA+ changes only from 44.61/18.43 to 44.10/16.50, suggesting that dummy prompts have a negligible effect.

\section{Implementation and Training Details}
\label{sec:implementation}

\subsection{Implementation}
\begin{figure}
    \centering
    \includegraphics[width=1\linewidth]{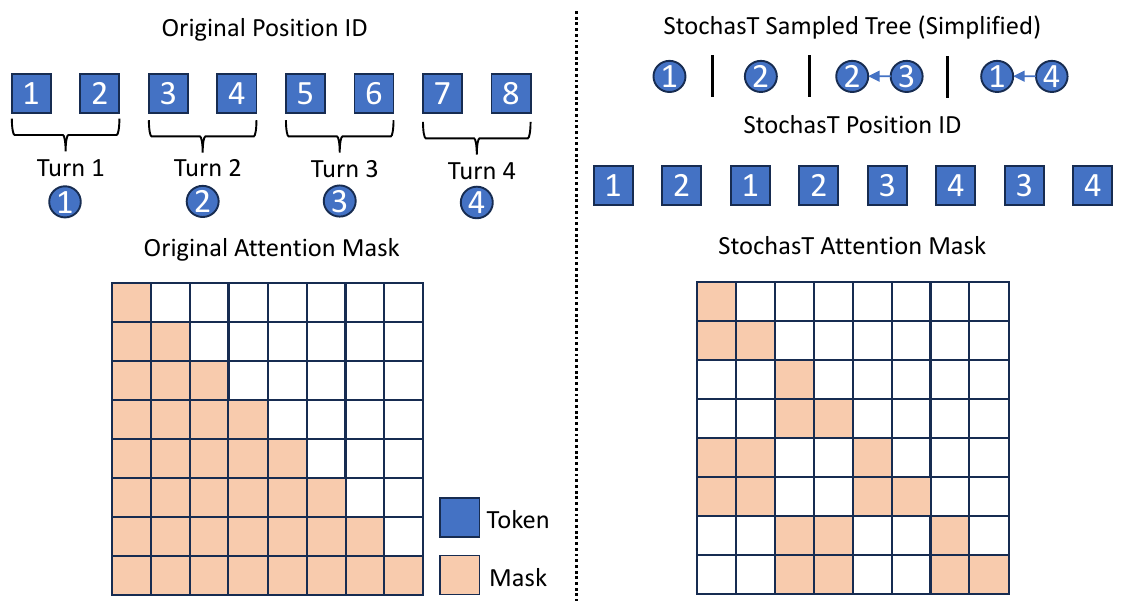}
    \caption{\textbf{Implementation}: Attention mask and position ID.}
    \label{fig:attention}
\end{figure}

We illustrate the implementation of attention masks and position IDs in StochasT in~\cref{fig:attention} using a 4-turn example, where each turn is simplified to 2 tokens for clarity. After sampling, turns 1 and 2 are connected to the root, so their token position IDs are both set to 1 and 2, while the position IDs of turns 3 and 4 are reset accordingly. The attention mask is updated for each token based on the sampled tree structure. This does not affect KV caching, as StochasT is applied only during training.

\subsection{Dataset Details}

\begin{figure}[h]
    \centering
    \includegraphics[width=1\linewidth]{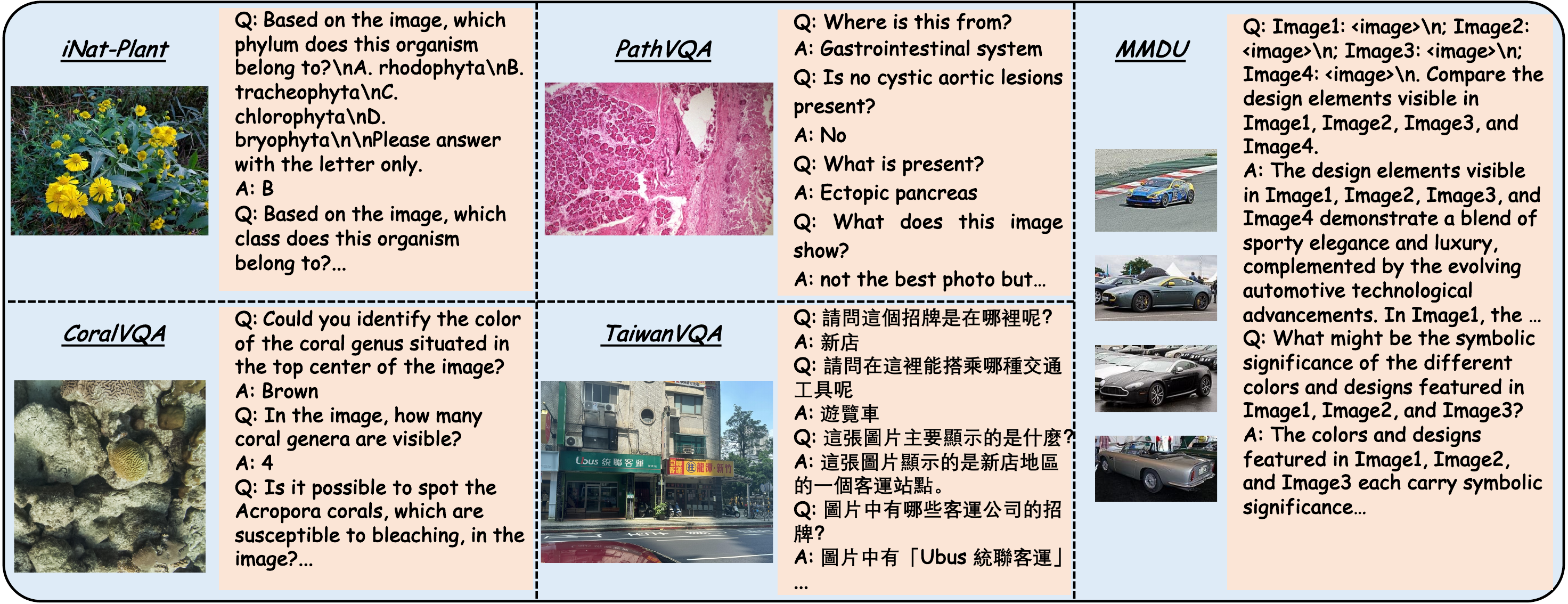}
    \caption{Examples of visual instruction tuning data sampled from each of the five downstream tasks.}
    \label{fig:examples}
\end{figure}

We summarize the comprehensive statistics of the datasets utilized for both training (\cref{tab:dataset_stats}) and evaluation (\cref{tab:dataset_stats_eval}), specifically detailing the total number of unique images and the average number of conversational turns per sample. Furthermore, \cref{fig:examples} illustrates representative multi-turn data samples drawn from each evaluated domain: iNat-Plant~\cite{tan2025visionllmsbadhierarchical}, PathVQA~\cite{he2020pathvqa}, CoralVQA~\cite{han2025coralvqa}, TaiwanVQA~\cite{hsieh2025taiwanvqa}, and MMDU~\cite{liu2024mmdu}.

\begin{table}[ht]
\centering
\caption{Statistics of the datasets used for training.}
\label{tab:dataset_stats}
\begin{tabular}{lccc}
\toprule
\textbf{Dataset} & \textbf{\# Images} & \textbf{\# Training Samples} & \textbf{Avg. Turns / Sample} \\
\midrule
PathVQA     & 2,599  & 2,599  & 6.92 \\
CoralVQA    & 10,536 & 10,536 & 7.98 \\
iNat-Plant  & 42,710 & 42,710 & 6.00 \\
TaiwanVQA   & 1,736  & 1,736  & 9.00 \\
MMDU        & 15,548 & 10,000 & 8.15 \\
\bottomrule
\end{tabular}
\end{table}

\begin{table}[ht]
\centering
\caption{Statistics of the datasets used for evaluation.}
\label{tab:dataset_stats_eval}
\begin{tabular}{lccc}
\toprule
\textbf{Dataset} & \textbf{\# Images} & \textbf{\# Eval. Instructions} & \textbf{Avg. Turns / Sample} \\
\midrule
PathVQA     & 831   & 5,723  & 6.89 \\
CoralVQA    & 487   & 3,896  & 8.00 \\
iNat-Plant  & 8,542 & 51,252 & 6.00 \\
TaiwanVQA   & 1,000 & 2,000  & 2.00 \\
MMDU        & 110   & 881    & 8.00 \\
\bottomrule
\end{tabular}
\end{table}

\subsection{Training Details}

\begin{table}[!h]
\centering
\caption{Training configurations for each dataset and task. All models are fine-tuned using LoRA with a warmup ratio of 0.03.}
\label{tab:training_config_combined}
\begin{tabular}{llccc}
\toprule
\textbf{Model} & \textbf{Task} & \textbf{Learning Rate} & \textbf{LoRA Rank} & \textbf{LoRA} $\boldsymbol{\alpha}$ \\
\midrule

\multirow{5}{*}{LLaVA-1.5-7B}
& PathVQA    & $1\text{e-}4$ & 64  & 64  \\
& CoralVQA   & $1\text{e-}4$ & 64  & 64  \\
& iNat-Plant & $1\text{e-}4$ & 64  & 64  \\
& TaiwanVQA  & $1\text{e-}4$ & 16  & 16  \\
& MMDU       & $2\text{e-}4$ & 256 & 256 \\
\midrule

\multirow{5}{*}{Qwen2.5-VL-3B}
& PathVQA    & $5\text{e-}4$ & 64  & 64  \\
& CoralVQA   & $1\text{e-}4$ & 64  & 64  \\
& iNat-Plant & $1\text{e-}4$ & 32  & 32  \\
& TaiwanVQA  & $1\text{e-}4$ & 32  & 32  \\
& MMDU       & $1\text{e-}4$ & 256 & 256 \\

\bottomrule
\end{tabular}
\end{table}

For our experiments, we utilize LoRA~\cite{hu2022lora} for parameter-efficient fine-tuning. All models are trained using two NVIDIA L40S, RTX A6000, or RTX Pro 6000 Blackwell GPUs. \Cref{tab:training_config_combined} summarizes the specific training configurations for each downstream task, detailing the learning rate, LoRA rank, and LoRA $\alpha$. Across all tasks, we employ a global batch size of 128 and a warmup ratio of 0.03. We train the models for multiple epochs and apply early stopping to select the checkpoint with the highest validation performance. Finally, we initialize our models using the instruction-tuned weights from \texttt{Qwen/Qwen2.5-VL-3B-Instruct} and \texttt{liuhaotian/llava-v1.5-7b}, available via HuggingFace.

\section{Additional Experiments and Results}
\label{sec:additional_exp}

\subsection{Comparison to Turn Dropout}
\label{sec:turn_dropout}

While we draw inspiration upon Dropout~\cite{srivastava2014dropout} and Stochastic Depth~\cite{huang2016deep} when designing the approach in the main paper, we choose to keep all turns/loss tokens to maximize the utility of the training data. In this section, we discuss an alternative, more straightforward approach, called Turn Dropout (TD).

In this variant, we stochastically drop individual turns from the conversation based on a drop probability sampled from the Beta distribution. As demonstrated in \cref{tab:single_multi} (which reports performance after 1,000 training steps on the iNat-Plant dataset), this explicit dropping mechanism remains highly effective. TD consistently surpasses both the deterministic SingleT and MultiT baselines, achieving performance that is highly competitive with our complete StochasT method. Consequently, vanilla Turn Dropout serves as a viable and highly efficient alternative to our primary approach, particularly in scenarios where computational resources or memory constraints are the primary bottleneck.

\begin{table}[h]
\centering
\caption{Performance comparison of different training paradigms under SingleT and MultiT evaluation settings on iNat-Plant (evaluated at 1,000 steps). Best results are \textbf{bolded}, and second-best are \underline{underlined}.}
\label{tab:single_multi}
\begin{tabular}{lcc}
\toprule
\textbf{Method} & \textbf{SingleT Eval} & \textbf{MultiT Eval} \\
\midrule
MultiT   & 84.74 & 91.12 \\
SingleT  & 88.84 & 88.81 \\
TD       & \underline{89.97} & \underline{91.66} \\
StochasT & \textbf{91.48} & \textbf{92.03} \\
\bottomrule
\end{tabular}
\end{table}

\subsection{Full Fine-Tuning on LLaVA-150K}
\label{sec:full_finetuning}

Because our proposed method is entirely data- and model-agnostic, we also validate its efficacy during the general visual instruction tuning stage immediately following model pre-training. For this experiment, we utilize the comprehensive LLaVA-150K dataset~\cite{liu2023visual}. We initialize the vision-language connector using the official pre-trained projector weights\footnote{HuggingFace: \texttt{liuhaotian/llava-v1.5-mlp2x-336px-pretrain-vicuna-7b-v1.5}} and the base LLM, Vicuna-7B~\cite{vicuna2023}, from the HuggingFace repository: \texttt{lmsys/vicuna-7b-v1.5}.

Following the standard LLaVA-1.5~\cite{Liu_2024_CVPR} training configuration, we train our model using full fine-tuning. Training is performed on four 96GB RTX 6000 Blackwell GPUs for a single epoch, employing a learning rate of $2\text{e-}5$, a warmup ratio of 0.03, and a global batch size of 128. 

We compare StochasT against the standard MultiT fine-tuning baseline utilized by LLaVA, evaluating both models on the MME benchmark~\cite{fu2025mme}. As detailed in \cref{tab:mme}, our method systematically outperforms the MultiT baseline across every single category. Notably, StochasT achieves a relative improvement of 19.36\% in overall perception capability and 25.56\% in reasoning capability. This demonstrates that applying StochasT during the foundational instruction tuning stage not only fortifies contextual robustness for downstream tasks but also amplifies the model's generalized multimodal capabilities.

\begin{table}[h]
\centering
\caption{Performance comparison of our method against the standard MultiT baseline on the MME benchmark.}
\label{tab:mme}
\begin{tabular}{lrr}
\toprule
\textbf{Category} & \textbf{MultiT} & \textbf{StochasT} \\ \midrule
OCR & 72.50 & \textbf{80.00} \\
Artwork & 71.00 & \textbf{83.25} \\
Celebrity & 68.24 & \textbf{102.06} \\
Code Reasoning & 57.50 & \textbf{80.00} \\
Color & 80.00 & \textbf{105.00} \\
Commonsense Reasoning & 99.29 & \textbf{102.14} \\
Count & 100.00 & \textbf{146.67} \\
Existence & 158.33 & \textbf{180.00} \\
Landmark & 108.75 & \textbf{120.75} \\
Numerical Calculation & 50.00 & \textbf{57.50} \\
Position & 46.67 & \textbf{50.00} \\
Posters & 87.76 & \textbf{102.38} \\
Scene & 146.25 & \textbf{151.25} \\
Text Translation & 0.00 & \textbf{20.00} \\ \midrule
\textbf{Perception} & 939.49 & \textbf{1121.36} \\
\textbf{Reasoning} & 206.79 & \textbf{259.64} \\
\bottomrule
\end{tabular}
\end{table}

\subsection{Budget-Controlled Scaling Experiment on a Newer Model}
\label{sec:new_models}

To further demonstrate the advantage of StochasT over SingleT and evaluate its effectiveness on a larger and more up-to-date model, we conduct budget-controlled experiments on iNat-Plant using Qwen3VL-32B~\cite{bai2025qwen3vltechnicalreport}. Specifically, all training settings are matched to use the same number of training tokens. As shown in~\cref{tab:new_model}, StochasT still substantially outperforms both MultiT and SingleT, improving by around 3 CRA points and 5 CRA+ points. These results suggest that StochasT remains effective under model scaling and continues to mitigate visual attention decay and contextual overfitting even with a stronger backbone.

\begin{table}[h]
    
    \centering
    \caption{Budget-controlled experiment on Qwen3VL-32B~\cite{bai2025qwen3vltechnicalreport} on iNat-Plant.}
    \begin{tabular}{l | ccc}
    \toprule
    \multirow{2}{*}{Metric} 
    & \multicolumn{3}{c}{Qwen3VL-32B~\cite{bai2025qwen3vltechnicalreport}} \\
    \cmidrule(lr){2-4}
    & MultiT & SingleT & \textbf{Ours} \\
    \midrule
    CRA  & 91.67 & 91.31 & \textbf{94.53} \\
    CRA+ & 83.31 & 84.38 & \textbf{88.16} \\
    \bottomrule
    \end{tabular}
    \label{tab:new_model}
    \vspace{-12pt}
\end{table}

\subsection{Per-dataset Results}

We detail the per-dataset evaluation accuracies in \cref{tab:main}, and report the Balanced Latin Square evaluation metrics: Context-Robust Accuracy (CRA) and Strict Contextual Consistency (CRA+), in \cref{tab:sft}. Our proposed StochasT method consistently outperforms standard MultiT training across all tasks. Furthermore, it achieves superior or highly comparable performance to SingleT training in the vast majority of settings, demonstrating exceptional versatility across all evaluation setups.
 
\begin{table}[h]
\vspace{-12pt}
\centering
\caption{Performance comparison of MultiT and SingleT training baselines against our proposed StochasT under both SingleT and MultiT evaluation protocols. Best results are \textbf{bolded}, and second-best are \underline{underlined}.}
\label{tab:main}

\setlength{\tabcolsep}{1.8pt}
\renewcommand{\arraystretch}{1.1}

\newcolumntype{C}{>{\centering\arraybackslash}p{0.92cm}}

\begin{tabular}{l *{8}{C} c}
\toprule

\multirow{2}{*}{\textbf{Method}}
& \multicolumn{2}{c}{\textbf{iNat-Plant}}
& \multicolumn{2}{c}{\textbf{PathVQA}}
& \multicolumn{2}{c}{\textbf{CoralVQA}}
& \multicolumn{2}{c}{\textbf{TaiwanVQA}}
& \textbf{MMDU} \\

\cmidrule(lr){2-3}
\cmidrule(lr){4-5}
\cmidrule(lr){6-7}
\cmidrule(lr){8-9}
\cmidrule(lr){10-10}

& SingleT & MultiT
& SingleT & MultiT
& SingleT & MultiT
& SingleT & MultiT
& MultiT \\

\midrule
LLaVA-1.5-7B                   & 59.60 & 61.44 & 33.83 & 39.98 & 36.31 & 41.03 & 45.20 & 44.9 & 10.81 \\
\quad +MultiT                  & 79.24 & \underline{88.95} & 41.64 & 61.72 & \textbf{79.15} & \underline{78.02} & 46.0 & 45.4 & \underline{12.68} \\
\quad +SingleT                 & \underline{85.89} & 83.11 & \textbf{64.41} & \textbf{67.05} & 76.44 & 77.54 & \textbf{47.1} & \textbf{46.7} & N/A \\
\quad +StochasT                & \textbf{88.14} & \textbf{89.07} & \underline{54.48} & \underline{66.36} & \textbf{79.15} & \textbf{80.88} & \underline{46.85} & \underline{46.20} & \textbf{13.10} \\
\midrule

Qwen2.5-VL-3B                  & 69.29 & 69.46 & 35.70 & 43.68 & 42.74 & 43.76 & 74.20 & 75.75 & 47.0 \\
\quad +MultiT                  & 86.10 & \underline{91.98} & 43.84 & 62.55 & 79.54 & 78.87 & \textbf{75.30} & \underline{76.15} & \underline{57.9} \\
\quad +SingleT                 & \textbf{91.93} & 91.16 & \textbf{62.48} & \underline{66.38} & \textbf{80.93} & \textbf{81.47} & 71.50 & 72.00 & N/A \\
\quad +StochasT                & \underline{91.78} & \textbf{92.68} & \underline{61.79} & \textbf{70.54} & \underline{80.23} & \underline{80.93} & \textbf{75.30} & \textbf{76.50} & \textbf{59.8} \\

\bottomrule
\end{tabular}
\end{table}

\begin{table}[h]
\centering
\caption{Robustness evaluation comparing Context-Robust Accuracy (CRA) and Strict Contextual Consistency (CRA+) across different training paradigms. Best results are \textbf{bolded}, and second-best are \underline{underlined}.}
\label{tab:sft}

\renewcommand{\arraystretch}{1.1}

\newcolumntype{C}{>{\centering\arraybackslash}p{0.92cm}}

\begin{tabular}{l*{8}{C}}
\toprule

\multirow{2}{*}{\textbf{Method}}
& \multicolumn{2}{c}{\textbf{iNat-Plant}}
& \multicolumn{2}{c}{\textbf{PathVQA}}
& \multicolumn{2}{c}{\textbf{CoralVQA}}
& \multicolumn{2}{c}{\textbf{TaiwanVQA}} \\

\cmidrule(lr){2-3}
\cmidrule(lr){4-5}
\cmidrule(lr){6-7}
\cmidrule(lr){8-9}

& CRA & CRA+
& CRA & CRA+
& CRA & CRA+
& CRA & CRA+ \\

\midrule

LLaVA-1.5-7B                   & 59.70 & 36.88 & 39.15 & 15.46 & 38.52 & 21.93 & 45.53 & 36.35 \\
\quad +MultiT                  & 72.54 & 62.37 & 56.94 & 6.67 & \underline{72.07} & 59.12 & 45.73 & 36.60 \\
\quad +SingleT                 & \underline{81.85} & \underline{72.24} & \textbf{68.41} & \textbf{40.92} & 71.77 & \underline{60.67} & \textbf{47.30} & \underline{38.75} \\
\quad +DM-SFT                  & \textbf{87.28} & \textbf{76.04} & \underline{63.85} & \underline{24.11} & \textbf{73.09} & \textbf{64.30} & \underline{46.55} & \textbf{39.10} \\
\midrule

Qwen2.5-VL-3B                  & 69.03 & 50.50 & 43.65 & 13.16 & 43.13 & 21.42 & 73.22 & 68.75 \\
\quad +MultiT                  & 87.72 & 72.06 & 59.58 & 16.04 & 71.88 & 57.99 & \underline{75.35} & \underline{71.10} \\
\quad +SingleT                 & \textbf{92.13} & \textbf{84.18} & \textbf{70.75} & \textbf{43.88} & \textbf{74.34} & \underline{62.81} & 71.80 & 67.40 \\
\quad +DM-SFT                  & \underline{92.12} & \underline{84.13} & \underline{65.37} & \underline{28.08} & \underline{73.04} & \textbf{63.32} & \textbf{75.58} & \textbf{71.50} \\

\bottomrule
\end{tabular}
\end{table}

\section{Qualitative Examples}
\label{sec:qualitative}

\begin{figure}[ht]
    \centering
    \includegraphics[width=1\linewidth]{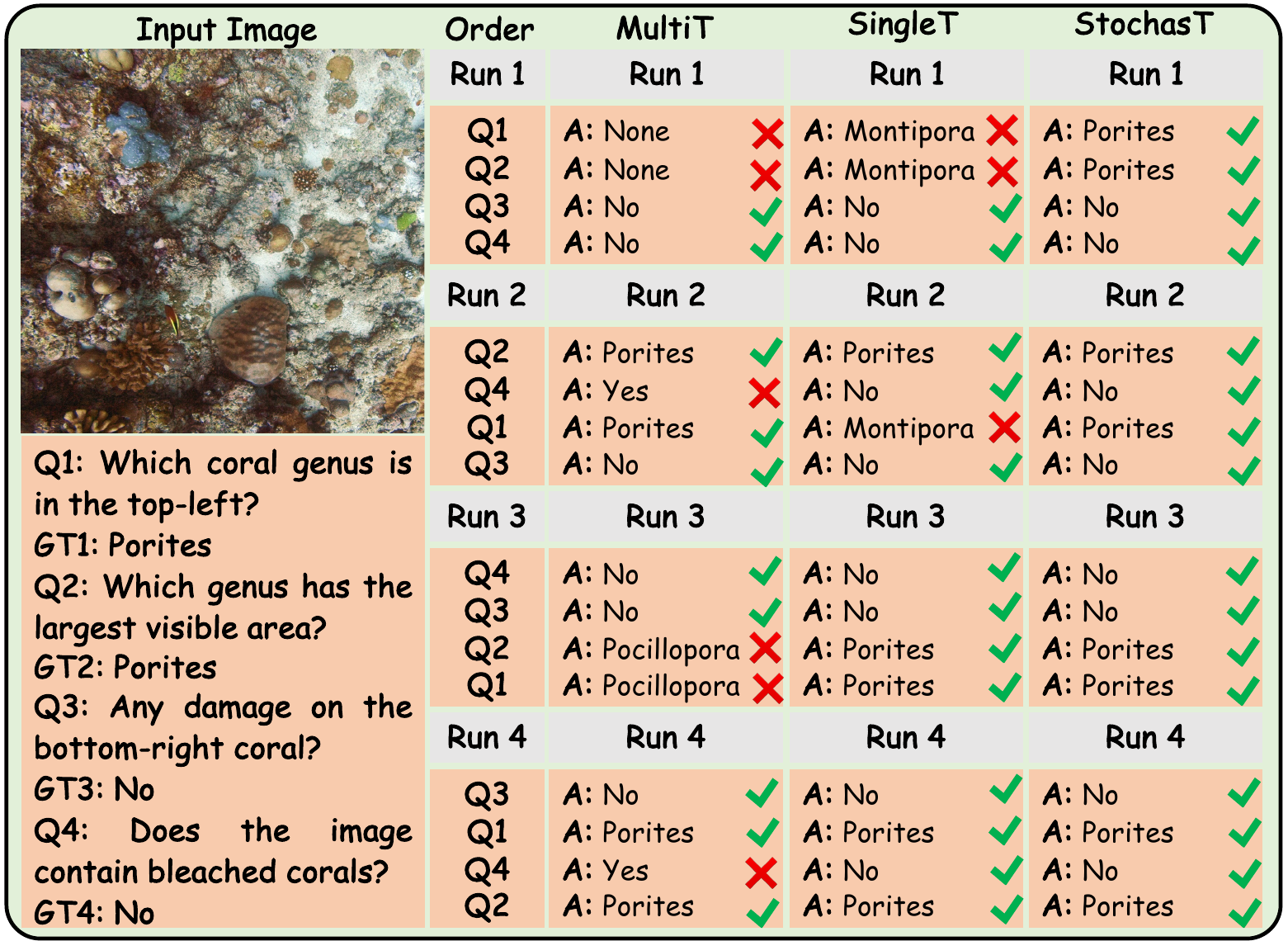}
    \caption{\textbf{Qualitative comparison of model robustness on CoralVQA using the Balanced Latin Square evaluation framework.} While the MultiT and SingleT baselines exhibit inconsistent predictions and high sensitivity to question ordering across the permutations (Runs 1–4), our proposed StochasT maintains 100\% accuracy and consistency. This visualization clearly demonstrates StochasT's capacity to generate reliable, context-invariant outputs.}
    \label{fig:qualitative_example}
\end{figure}

We present a qualitative example from our Balanced Latin Square evaluation to compare StochasT against the MultiT and SingleT baselines on the CoralVQA dataset. As illustrated in \cref{fig:qualitative_example}, applying StochasT yields robust, context-invariant predictions that remain entirely stable under diverse sequence permutations. This consistent behavior perfectly aligns with the strict reliability demands required for practical, real-world deployment scenarios.

\end{document}